\documentclass[conference]{IEEEtran}
\IEEEoverridecommandlockouts
\usepackage{cite}
\usepackage[numbers,sort&compress]{natbib}

\usepackage[utf8]{inputenc}
\usepackage{amsmath,amssymb}
\usepackage{graphicx}
\usepackage{bm}
\usepackage{enumerate}
\usepackage{comment}
\usepackage{xcolor}
\usepackage{url}
\usepackage{color,soul}
\usepackage{subcaption}
\usepackage{float}

\definecolor{light-gray}{gray}{0.8}

\usepackage{amsmath,amssymb,amsfonts}
\usepackage{algorithmic}
\usepackage[linesnumbered,ruled]{algorithm2e}
\usepackage{graphicx}
\usepackage{textcomp}
\usepackage{xcolor}
\usepackage{subcaption}

\def\BibTeX{{\rm B\kern-.05em{\sc i\kern-.025em b}\kern-.08em
    T\kern-.1667em\lower.7ex\hbox{E}\kern-.125emX}}

\makeatletter
\newcommand{\linebreakand}{%
  \end{@IEEEauthorhalign}
  \hfill\mbox{}\par
  \mbox{}\hfill\begin{@IEEEauthorhalign}
}
\makeatother

\begin{document}

\title{Leveraging Large Language Model for Intelligent Log Processing and Autonomous Debugging in Cloud AI Platforms\\}

\author{

\small % Set font size to 10pt

\begin{tabular}[t]{c@{\extracolsep{8em}}c} 

\textsuperscript{} Cheng Ji & Huaiying Luo\\
\textsuperscript{} Siebel School of Computing and Data Science & College of Computing and Information Science \\
\textit{University of Illinois at Urbana-Champaign} & \textit{Cornell University}\\
\textsuperscript{}Champaign, Illinois, USA & New York, USA \\
\textsuperscript{}chengji5@illinois.edu & hl2446@cornell.edu \\

\\

\end{tabular}
}

\maketitle

\begin{abstract}
With the increasing complexity and rapid expansion of the scale of AI systems in cloud platforms, the log data generated during system operation is massive, unstructured, and semantically ambiguous, which brings great challenges to fault location and system self-repair. In order to solve this problem, this paper proposes an intelligent log processing and automatic debugging framework based on Large Language Model (LLM), named Intelligent Debugger (LLM-ID). This method is extended on the basis of the existing pre-trained Transformer model, and integrates a multi-stage semantic inference mechanism to realize the context understanding of system logs and the automatic reconstruction of fault chains. Firstly, the system log is dynamically structured, and the unsupervised clustering and embedding mechanism is used to extract the event template and semantic schema. Subsequently, the fine-tuned LLM combined with the multi-round attention mechanism to perform contextual reasoning on the log sequence to generate potential fault assumptions and root cause paths. Furthermore, this paper introduces a reinforcement learning-based policy-guided recovery planner, which is driven by the remediation strategy generated by LLM to support dynamic decision-making and adaptive debugging in the cloud environment. Compared with the existing rule engine or traditional log analysis system, the proposed model has stronger semantic understanding ability, continuous learning ability and heterogeneous environment adaptability. Experiments on the cloud platform log dataset show that LLM-ID improves the fault location accuracy by 16.2\%, which is significantly better than the current mainstream methods
\end{abstract}

\begin{IEEEkeywords}
Large language models, Log analysis, Automatic commissioning, Cloud platform, Failure recovery
\end{IEEEkeywords}

\section{Introduction}
With the rapid development of cloud computing and artificial intelligence (AI) technology, the deployment of modern AI systems on cloud platforms has become mainstream. These systems are typically composed of multiple heterogeneous components that run in a distributed environment and generate massive amounts of log data \cite{yang2024hades}. These logs not only record the operating status, error information, and performance metrics of the system, but also contain various information such as user behavior data, system configuration, and resource usage.

The diversity and complexity of logs make them an integral part of cloud platform AI systems. As the system scales, the amount of log data generated grows exponentially, further increasing the difficulty of log analysis. For system maintenance personnel, these logs are an important basis for locating problems, troubleshooting, and optimizing system performance \cite{sanodia2024revolutionizing, jin2025scalability}. Through in-depth analysis of log data, potential problems in the system can be quickly identified, fault prediction can be carried out, and corresponding repair measures can be taken to ensure the stability and efficient operation of the system.

However, traditional log analysis methods are inadequate in the face of large-scale, heterogeneous, and dynamically changing log data. Manually analyzing logs is not only inefficient, but also prone to missing critical information, leading to delays in fault location and repair \cite{pissanidis2023integrating}. For example, manual analysis often relies on keyword searches, fixed rules, and empirical judgments, making it difficult to handle high-dimensional log data and cope with semantic changes and complexity in log content. In addition, the existing log analysis tools mostly rely on static rules and manual intervention, and lack the ability to deeply understand and automate the behavior of complex systems. When multiple failures occur, traditional tools often struggle to work together, resulting in a complex and error-prone diagnostic process \cite{li2024advances}. As a result, the limitations of traditional approaches make efficient log analysis and troubleshooting in large-scale cloud platforms challenging. 

In recent years, with the rise of large language models (LLMs) and generative AI technologies \cite{mahesh2020machine,grattafiori2024llama, li2024vqa, xu2024style, goodfellow2014generative, pinheiro2021variational, zhang2022covid, 10.1145/3627673.3679071}, researchers have begun to explore the application of these advanced technologies in diverse fields like finance, social network, linguistics, biology and healthcare\cite{jordan2015machine, sui2024ensemble, zhao2024hedge, yang2024comparative, li2024deception, ji-etal-2024-rag, 10679029, zhong2025narrative}. Consequently, it is reasonable to investigate their potential in the domain of log analysis and fault debugging. With its powerful language understanding, reasoning and generation capabilities \cite{achiam2023gpt, he2025givestructuredreasoninglarge,he2025selfgiveassociativethinkinglimited, wang2024enhancing, 10628639}, LLMs can extract semantic information from massive logs, identify potential anomalous patterns, generate fault diagnosis reports, and provide remediation suggestions \cite{yang2025research}. Through in-depth analysis of the context of the logs, the LLM can identify the implicit associations and causal relationships in the logs, and can not only find anomalies in the data, but also propose fixes through generative models. Furthermore, it can combine with computer vision methods to handle non-text information \cite{ li2025visual, li2024segmentation}. These technological breakthroughs make log analysis not only limited to simple rule matching, but also develop in a more intelligent and automated direction \cite{jin2025adaptive}. For example, generative AI can automatically generate possible root causes of failures based on historical fault logs and predict potential system bottlenecks in the future, greatly improving the accuracy and response speed of fault predictions.

However, there are still many challenges in applying LLMs to log analysis and automatic debugging of cloud platform AI systems. First, the diversity and complexity of log data make the model training and inference process highly complex \cite{liu2024mt2st}. Different types of logs often have different formats and structures, and missing data, exceptions, and noise are common problems, which make the model require additional pre-processing work when processing.

\section{Related Work}
Xu et al. \cite{xu2024empowering} proposed a framework for integrating artificial intelligence (AI) technologies into cloud services, aiming to empower software engineers with development capabilities. The framework helps developers develop and maintain software more efficiently in the cloud environment by providing intelligent tools and services. Hrusto et al. \cite{hrusto2024autonomous} proposed an autonomous monitoring system for early detection of failures and reporting interpretable alerts in cloud operations. The system combines performance metrics and log analysis, employing GPT-3 models for anomaly detection, capable of generating alerts with information on potential root causes.

Furthermore, Stutz et al. \cite{stutz2024enhancing} use machine learning and deep learning techniques to monitor and respond to security threats in cloud environments in real time. The architecture improves the security of cloud systems by detecting and preventing various security threats such as data breaches, malware attacks, and unauthorized access. Hassan \cite{hassan2025managing} analyzes historical execution logs, and the model can predict potential bottlenecks and points of failure, enabling more efficient resource scheduling and performance optimization.

Tupe and Thube \cite{tupe2025ai} propose an enterprise API architecture adaptation for AI agent workflows. With the rapid development of generative AI, the traditional enterprise API architecture mainly focuses on predefined interaction patterns driven by humans, which is difficult to support the dynamic goal-oriented behavior of intelligent agents. Chen et al. \cite{chen2024transforming} proposed a hybrid cloud architecture transformation approach for emerging AI workloads through an innovative full-stack collaborative design approach that emphasizes availability, manageability, economics, adaptability, efficiency, and scalability.

Cheng et al. \cite{cheng2023logai} proposed LogAI, an open-source library for log analysis and intelligence, which aims to provide developers with a unified log analysis tool. LogAI supports a variety of AI-driven log analysis tasks, including log summarization, log clustering, and log anomaly detection. Tadi \cite{tadi2022architecting} notes that logs are a key source of data for troubleshooting, security audits, and performance monitoring.

\section{METHODOLOGIES}
\subsection{Multi-scale log structure abstraction and semantic inference}
The diversity of logging systems makes the original log essentially composed of unstructured, semantically repetitive, but heterogeneous expressions. In order to extract stable semantic patterns and avoid noise, we construct multi-scale windows and introduce a fuzzy-matching attention module (FAM) for event template abstraction. Specifically, we first construct a collection of local contexts using the window extractor, as shown in Equation 1:
\begin{equation}
\mathcal{W}^{(s)} = \left\{ \mathcal{W}_i^{(s)} = \left\{ e_i, \ldots, e_{i+s-1} \right\} \,\middle|\, i = 1, \ldots, T - s + 1 \right\}, \; s \in \mathcal{S}, \tag{1}
\end{equation}
where $\mathcal{S} = \{3, 5, 7\}$ is a multiscale window set. The similarity of each window to the prototype template vector $p_k$ is calculated by the fuzzy matching attention score, as shown in Equations (2) and (3):

\begin{equation}
\alpha_{i,k}^{(s)} = \frac{\exp\left( - \frac{ \left\| \bar{w}_i^{(s)} - p_k \right\|^2 }{ \tau } \right)}{ \sum_j \exp\left( - \frac{ \left\| \bar{w}_i^{(s)} - p_k \right\|^2 }{ \tau } \right) }, \tag{2}
\end{equation}

\begin{equation}
\bar{w}_i^{(s)} = \frac{1}{s} \sum_{j=0}^{s-1} e_{i+j}. \tag{3}
\end{equation}

Finally, the multi-scale clustering weighting results are synthesized to construct a unified event template embedding $t_i$, as shown in Equation (4):

\begin{equation}
t_i = \sum_{s \in \mathcal{S}} \sum_k \alpha_{i,k}^{(s)} \cdot p_k. \tag{4}
\end{equation}

In this way, we map the original log sequence $\mathcal{L}$ to the structurally stable event sequence $\mathcal{T} = \{t_1, \ldots, t_T\}$ eliminating redundancy, ambiguity, and abnormal expressions, and laying the foundation for semantic inference. In the semantic inference phase, the goal is to mine the underlying causal chain from the sequence of events and predict possible points of failure. We use the Hierarchical Multi-Hop Attention Reasoning mechanism to integrate the reasoning ability of LLM with the graph structure construction. Let the semantic representation of the Transformer output layer $l$ be $H^{(l)}$, and we define the bidirectional attention graph between events as Equations 5 and 6:

\begin{equation}
A_{ij} = \frac{\exp\left( \mathrm{sim}(h_i, h_j) \right)}{\sum_k \exp\left( \mathrm{sim}(h_i, h_k) \right)}, \tag{5}
\end{equation}

\begin{equation}
\mathrm{sim}\left(h_i, h_j\right) = \frac{h_i^\top W_a h_j}{\left\| h_i \right\| \cdot \left\| h_j \right\|}. \tag{6}
\end{equation}

In Equations 5 and 6, denotes the semantic representation vector of event at the layer, and represents the attention matrix that encodes the directed influence between events. The matrix is computed using scaled dot-product attention between semantic vectors. This graph structure captures temporal and semantic dependencies across events, forming the basis for multi-hop reasoning. In multiple rounds of inference, each round we update the node as Equation 7:

\begin{equation}
h_i^{(r+1)} = \mathrm{ReLU} \left( \sum_{j=1}^{T} A_{ij}^{(r)} \cdot h_j^{(r)} \right), \tag{7}
\end{equation}

where the initial $h_i^{(0)} = t_i$, a total of $R$ rounds (e.g., $R = 3$) are inferred, and the final root cause attention score is obtained, as shown in Equation 8:

\begin{equation}
\psi_i = \sigma \left( w_\psi^\top h_i^{(R)} \right), \quad \psi_i \in [0, 1]. \tag{8}
\end{equation}

In Equation 8, denotes the final root cause attention score for event , aggregated from rounds of bidirectional attention reasoning. The scalar reflects the accumulated influence weight of each node in the inferred fault chain graph.

\subsection{Confidence-guided recovery strategy}
The goal of the recovery strategy planner is to generate an optimal sequence of operations $\mathcal{A} = \{a_1, a_2, \ldots\}$ based on the current state $s_t$ and the failure prediction $f$ to fix the exception at the lowest cost. We propose a Bayesian Policy Shaping (BPs) based on Bayesian confidence priors to reconstruct the dynamic credibility of the policy output. The reinforcement learning strategy adopts an actor-critic structure, with the strategy $\pi_\theta(a \mid s)$ and the value function $V_\omega(s)$. We model action confidence using a Beta distribution, as in Equations 9 and 10:

\begin{equation}
P\left( \text{conf}_a \mid s \right) \sim \mathrm{Beta}(\alpha_a, \beta_a), \tag{9}
\end{equation}

\begin{equation}
\alpha_a, \beta_a = \mathrm{MLP}_{\text{prior}}(s). \tag{10}
\end{equation}

The final strategy output is as Equation 11:

\begin{equation}
\pi^*(a \mid s) = \pi_\theta(a \mid s) \cdot \mathbb{E}_{\mathrm{Beta}(\alpha_a, \beta_a)}[x]. \tag{11}
\end{equation}

In the training process, we jointly optimize the failure classification loss $\mathcal{L}_{\text{fault}}$, the causal chain contrast loss $\mathcal{L}_{\text{causal}}$, and the recovery strategy loss $\mathcal{L}_{\text{RL}}$ to form a joint goal, as shown in Equation 12:

\begin{equation}
\mathcal{L}_{\text{total}} = \mathcal{L}_{\text{fault}} + \lambda_1 \cdot \mathcal{L}_{\text{causal}} + \lambda_2 \cdot \mathcal{L}_{\text{RL}} + \lambda_3 \cdot KL\left(P_{\text{conf}} \parallel P_{\text{uniform}}\right). \tag{12}
\end{equation}

The \textbf{KL} regex guarantees that the confidence evaluation is convergent, preventing overfitting or the strategy from falling into extremes. Figure 1 illustrates the overall flow of the proposed model LLM-ID framework. The framework consists of three main modules: a Structuring Encoder, a Semantic Reasoning LLM, and a Policy-guided Recovery Planner.

\begin{figure}[h!]
  \centering
    \includegraphics[width=0.9\linewidth, height=0.45\linewidth]{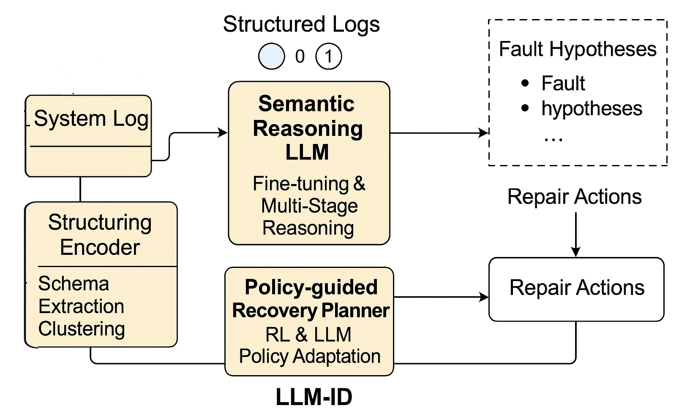}
    \caption{Framework of Proposed LLM-ID Model}
  \label{fig:bar}
\end{figure}

\section{EXPERIMENTS}
\subsection{Experimental setup}
The experiment uses the log dataset Loghub as the experimental dataset. Loghub contains more than 77GB of real-world system logs from distributed systems, operating systems, server applications, and more, covering millions of log records and supporting multiple log formats. The dataset has a high degree of diversity and authenticity, and a subset of the dataset also provides anomalous annotation information, which is suitable for supervised learning tasks.

All models were implemented in PyTorch and trained on an NVIDIA A100 80GB GPU server. For the LLM-ID framework, the structured encoder uses a Bi-LSTM with 2 layers and 256 hidden units. The LLM semantic reasoning module is based on a fine-tuned 6.7B parameter transformer with a context window size of 2048 tokens and rotary positional embeddings. The recovery planner employs an actor-critic algorithm with a learning rate of 5e-5, gamma = 0.99, entropy coefficient of 0.01, and batch size of 64. All models are trained for 50 epochs with early stopping based on validation loss. Each experiment is repeated 3 times, and the average results are reported. The log sequence length is fixed to 512 tokens.

In order to verify the effectiveness of the proposed LLM-ID framework, we selected four log analysis and fault debugging methods as comparison methods including:
\begin{itemize}
    \item Deformable DETR (Deformable Detection Transformer) can flexibly handle irregular and noisy log data by introducing a deformable attention mechanism.
    \item Graph Convolutional Networks for Fault Recovery (GCN-FR) Graph Convolutional Networks is used to build graphs of log events and capture correlations and dependencies between events using graph convolutional layers. GCN-FR leverages graph neural networks to model the interaction of time-series relationships and multi-dimensional data in logs to help achieve more accurate fault recovery strategies.
    \item The Transformer-based Fault Localization and Recovery (TL-FD/FR) Transformer captures the time series information in the logs and generates the root cause path of the fault through a multi-layer self-attention network. Combined with the enhanced Transformer model. 
    \item Semi-supervised Learning for Anomaly Detection (SSL-AD) performs anomaly detection by combining a small amount of fault annotated data with a large amount of unlabeled normal data. SSL-AD is able to effectively extract potentially anomalous patterns from large amounts of log data by building autoencoders and adversarial training methods.
\end{itemize}

\subsection{Experimental analysis}
Log analytics throughput (LAT) measures the amount of log data that a model is able to process per second. 
\begin{figure}[h!]
  \centering
    \includegraphics[width=0.9\linewidth, height=0.45\linewidth]{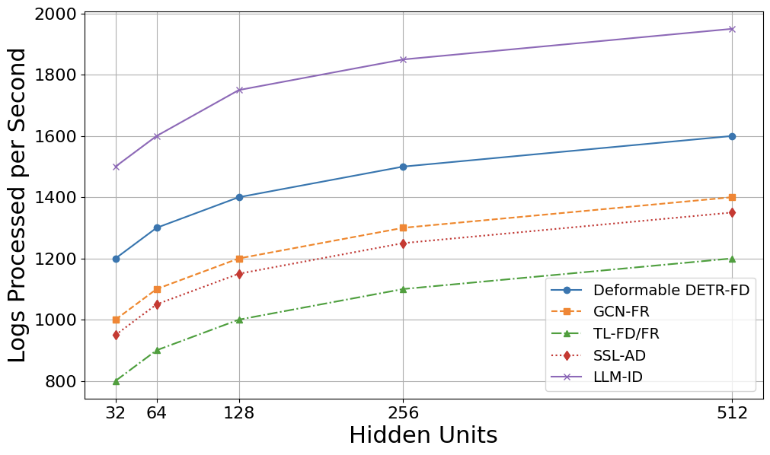}
    \caption{Log Analysis Throughput Comparison by Model Parameter}
  \label{fig:bar}
\end{figure}

This metric reflects the processing capacity of the system under large-scale log data, which can reflect the efficiency and scalability of the model. Figure 2 shows that with the increase of the number of hidden units, the log throughput of each method increases, but the increase is different. LLM-ID performs best in all parameter configurations, and its throughput increases from 1500 to 1950 records per second from 32 to 512 hidden cells, an increase of more than 30\%, indicating that it has good utilization efficiency for the expansion of model capacity.

Deformable DETR-FD followed closely behind, also showing a clear upward trend, but the increase (about 33\%) was slightly lower than that of LLM-ID, indicating that the deformable attention mechanism is beneficial to throughput. GCN-FR and SSL-AD grow faster in the range of medium and low parameters (from $\sim1000$ to $\sim1400$), but tend to be flat when they reach larger parameters, reflecting the limited sensitivity of graph convolution and semi-supervised methods to model size. TL-FD/FR has grown steadily despite having the lowest base throughput, validating the scalability of the Transformer architecture in log processing.

Debugging Recovery Time is a key performance indicator that measures the time it takes for a system to complete automatic or semi-automatic remediation from the detection of a failure. As the number of transformer layers increases from 2 to 10 in Figure 3, the failure recovery time of each method shows a steady downward trend, indicating that deeper models can help improve the accuracy and efficiency of log understanding and recovery decisions.

\begin{figure}[h!]
  \centering
    \includegraphics[width=0.9\linewidth, height=0.45\linewidth]{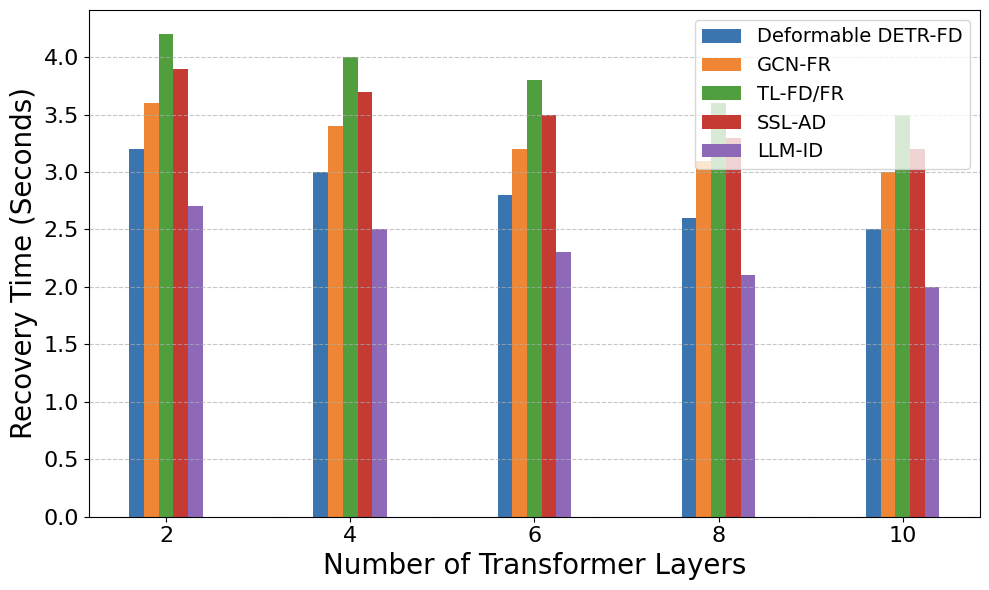}
    \caption{Debugging Recovery Time by Number of Transformer Layers}
  \label{fig:bar}
\end{figure}

Among them, LLM-ID maintains the lowest recovery delay (from 2.7 s to 2.0 s) at all layers, highlighting its advantages in adaptive debugging strategy and multi-stage semantic reasoning. Deformable DETR-FD followed closely behind, with the recovery time decreasing from 3.2 s to 2.5 s, indicating the help of the deformable attention mechanism for fine log pattern capture. GCN-FR and SSL-AD decreased from about 3.6 s/3.9 s to 3.0 s/3.2 s, respectively, reflecting the sensitivity of graph convolution and semi-supervised methods to layer expansion. However, the recovery time of TL-FD/FR is the highest (4.2 s→3.5 s), suggesting that pure Transformer is not inferior to other methods in terms of recovery speed.

\section{CONCLUSION}
In conclusion, this study constructs an LLM-ID framework combining deep multi-stage semantic inference and policy optimization to achieve faster and more stable fault recovery under multiple model parameter configurations: compared with traditional deformable attention, graph convolution, pure transformer, and semi-supervised methods, LLM-ID always maintains the lowest recovery delay, which fully proves its synergistic advantages in log understanding and adaptive debugging strategies. In the future, the combination of model inference and edge computing can be further explored to achieve real-time self-healing in distributed scenarios. At the same time, adversarial sample training and meta-learning mechanisms can be introduced to improve the robustness and generalization ability of new failure modes.

\renewcommand{\bibfont}{\footnotesize}

\footnotesize{
\bibliographystyle{IEEEtran}
\bibliography{main}

% Generated by IEEEtran.bst, version: 1.12 (2007/01/11)
\begin{thebibliography}{10}
\providecommand{\url}[1]{#1}
\csname url@samestyle\endcsname
\providecommand{\newblock}{\relax}
\providecommand{\bibinfo}[2]{#2}
\providecommand{\BIBentrySTDinterwordspacing}{\spaceskip=0pt\relax}
\providecommand{\BIBentryALTinterwordstretchfactor}{4}
\providecommand{\BIBentryALTinterwordspacing}{\spaceskip=\fontdimen2\font plus
\BIBentryALTinterwordstretchfactor\fontdimen3\font minus \fontdimen4\font\relax}
\providecommand{\BIBforeignlanguage}[2]{{%
\expandafter\ifx\csname l@#1\endcsname\relax
\typeout{** WARNING: IEEEtran.bst: No hyphenation pattern has been}%
\typeout{** loaded for the language `#1'. Using the pattern for}%
\typeout{** the default language instead.}%
\else
\language=\csname l@#1\endcsname
\fi
#2}}
\providecommand{\BIBdecl}{\relax}
\BIBdecl

\bibitem{yang2024hades}
Z.~Yang, Y.~Jin, and X.~Xu, ``Hades: Hardware accelerated decoding for efficient speculation in large language models,'' \emph{arXiv preprint arXiv:2412.19925}, 2024.

\bibitem{sanodia2024revolutionizing}
G.~Sanodia, ``Revolutionizing cloud modernization through ai integration,'' \emph{Turkish Journal of Computer and Mathematics Education}, vol.~15, no.~2, pp. 266--283, 2024.

\bibitem{jin2025scalability}
Y.~Jin and Z.~Yang, ``Scalability optimization in cloud-based ai inference services: Strategies for real-time load balancing and automated scaling,'' \emph{arXiv preprint arXiv:2504.15296}, 2025.

\bibitem{pissanidis2023integrating}
D.~L. Pissanidis and K.~Demertzis, ``Integrating ai/ml in cybersecurity: An analysis of open xdr technology and its application in intrusion detection and system log management,'' 2023.

\bibitem{li2024advances}
Z.~Li, S.~He, Z.~Yang, M.~Ryu, K.~Kim, and R.~Madduri, ``Advances in appfl: A comprehensive and extensible federated learning framework,'' \emph{arXiv preprint arXiv:2409.11585}, 2024.

\bibitem{mahesh2020machine}
B.~Mahesh \emph{et~al.}, ``Machine learning algorithms-a review,'' \emph{International Journal of Science and Research (IJSR).[Internet]}, vol.~9, no.~1, pp. 381--386, 2020.

\bibitem{grattafiori2024llama}
A.~Grattafiori, A.~Dubey, A.~Jauhri, A.~Pandey, A.~Kadian, A.~Al-Dahle, A.~Letman, A.~Mathur, A.~Schelten, A.~Vaughan \emph{et~al.}, ``The llama 3 herd of models,'' \emph{arXiv preprint arXiv:2407.21783}, 2024.

\bibitem{li2024vqa}
P.~Li, Q.~Yang, X.~Geng, W.~Zhou, Z.~Ding, and Y.~Nian, ``Exploring diverse methods in visual question answering,'' in \emph{2024 5th International Conference on Electronic Communication and Artificial Intelligence (ICECAI)}.\hskip 1em plus 0.5em minus 0.4em\relax IEEE, 2024, pp. 681--685.

\bibitem{xu2024style}
X.~Xu, Z.~Wang, Y.~Zhang, Y.~Liu, Z.~Wang, Z.~Xu, M.~Zhao, and H.~Luo, ``Style transfer: From stitching to neural networks,'' in \emph{2024 5th International Conference on Big Data \& Artificial Intelligence \& Software Engineering (ICBASE)}.\hskip 1em plus 0.5em minus 0.4em\relax IEEE, 2024, pp. 526--530.

\bibitem{goodfellow2014generative}
I.~J. Goodfellow, J.~Pouget-Abadie, M.~Mirza, B.~Xu, D.~Warde-Farley, S.~Ozair, A.~Courville, and Y.~Bengio, ``Generative adversarial nets,'' \emph{Advances in neural information processing systems}, vol.~27, 2014.

\bibitem{pinheiro2021variational}
L.~Pinheiro~Cinelli, M.~Ara{\'u}jo~Marins, E.~A. Barros~da Silva, and S.~Lima~Netto, ``Variational autoencoder,'' in \emph{Variational methods for machine learning with applications to deep networks}.\hskip 1em plus 0.5em minus 0.4em\relax Springer, 2021, pp. 111--149.

\bibitem{zhang2022covid}
T.~Zhang, B.~Zhang, F.~Zhao, and S.~Zhang, ``Covid-19 localization and recognition on chest radiographs based on yolov5 and efficientnet,'' in \emph{2022 7th International Conference on Intelligent Computing and Signal Processing (ICSP)}.\hskip 1em plus 0.5em minus 0.4em\relax IEEE, 2022, pp. 1827--1830.

\bibitem{10.1145/3627673.3679071}
H.~Xu, X.~Wang, and H.~Chen, ``Towards real-time and personalized code generation,'' in \emph{Proceedings of the 33rd ACM International Conference on Information and Knowledge Management}, 2024, p. 5568–5569.

\bibitem{jordan2015machine}
M.~I. Jordan and T.~M. Mitchell, ``Machine learning: Trends, perspectives, and prospects,'' \emph{Science}, vol. 349, no. 6245, pp. 255--260, 2015.

\bibitem{sui2024ensemble}
M.~Sui, C.~Zhang, L.~Zhou, S.~Liao, and C.~Wei, ``An ensemble approach to stock price prediction using deep learning and time series models,'' in \emph{2024 IEEE 6th International Conference on Power, Intelligent Computing and Systems (ICPICS)}.\hskip 1em plus 0.5em minus 0.4em\relax IEEE, 2024, pp. 793--797.

\bibitem{zhao2024hedge}
S.~Zhao, Z.~Dong, Z.~Cao, and R.~Douady, ``Hedge fund portfolio construction using polymodel theory and itransformer,'' \emph{arXiv preprint arXiv:2408.03320}, 2024.

\bibitem{yang2024comparative}
Q.~Yang, P.~Li, X.~Xu, Z.~Ding, W.~Zhou, and Y.~Nian, ``A comparative study on enhancing prediction in social network advertisement through data augmentation,'' in \emph{2024 4th International Conference on Machine Learning and Intelligent Systems Engineering (MLISE)}.\hskip 1em plus 0.5em minus 0.4em\relax IEEE, 2024, pp. 214--218.

\bibitem{li2024deception}
P.~Li, M.~Abouelenien, R.~Mihalcea, Z.~Ding, Q.~Yang, and Y.~Zhou, ``Deception detection from linguistic and physiological data streams using bimodal convolutional neural networks,'' in \emph{2024 5th International Conference on Information Science, Parallel and Distributed Systems (ISPDS)}.\hskip 1em plus 0.5em minus 0.4em\relax IEEE, 2024, pp. 263--267.

\bibitem{ji-etal-2024-rag}
\BIBentryALTinterwordspacing
Y.~Ji, Z.~Li, R.~Meng, S.~Sivarajkumar, Y.~Wang, Z.~Yu, H.~Ji, Y.~Han, H.~Zeng, and D.~He, ``{RAG}-{RLRC}-{L}ay{S}um at {B}io{L}ay{S}umm: Integrating retrieval-augmented generation and readability control for layman summarization of biomedical texts,'' in \emph{Proceedings of the 23rd Workshop on Biomedical Natural Language Processing}, D.~Demner-Fushman, S.~Ananiadou, M.~Miwa, K.~Roberts, and J.~Tsujii, Eds.\hskip 1em plus 0.5em minus 0.4em\relax Bangkok, Thailand: Association for Computational Linguistics, Aug. 2024, pp. 810--817. [Online]. Available: \url{https://aclanthology.org/2024.bionlp-1.75/}
\BIBentrySTDinterwordspacing

\bibitem{10679029}
Y.~Li, B.~Bohara, H.~S. Krishnamoorthy, and J.~Seshadrinath, ``Data-driven digital twins for monitoring the health and performance of converters,'' in \emph{2024 IEEE International Communications Energy Conference (INTELEC)}, 2024, pp. 1--6.

\bibitem{zhong2025narrative}
J.~Zhong, Y.~Wang, D.~Zhu, and Z.~Wang, ``A narrative review on large ai models in lung cancer screening, diagnosis, and treatment planning,'' \emph{arXiv preprint arXiv:2506.07236}, 2025.

\bibitem{achiam2023gpt}
J.~Achiam, S.~Adler, S.~Agarwal, L.~Ahmad, I.~Akkaya, F.~L. Aleman, D.~Almeida, J.~Altenschmidt, S.~Altman, S.~Anadkat \emph{et~al.}, ``Gpt-4 technical report,'' \emph{arXiv preprint arXiv:2303.08774}, 2023.

\bibitem{he2025givestructuredreasoninglarge}
\BIBentryALTinterwordspacing
J.~He, M.~D. Ma, J.~Fan, D.~Roth, W.~Wang, and A.~Ribeiro, ``Give: Structured reasoning of large language models with knowledge graph inspired veracity extrapolation,'' 2025. [Online]. Available: \url{https://arxiv.org/abs/2410.08475}
\BIBentrySTDinterwordspacing

\bibitem{he2025selfgiveassociativethinkinglimited}
\BIBentryALTinterwordspacing
J.~He, J.~Fan, B.~Jiang, I.~Houine, D.~Roth, and A.~Ribeiro, ``Self-give: Associative thinking from limited structured knowledge for enhanced large language model reasoning,'' 2025. [Online]. Available: \url{https://arxiv.org/abs/2505.15062}
\BIBentrySTDinterwordspacing

\bibitem{wang2024enhancing}
J.~Wang, Z.~Zhang, Y.~He, Y.~Song, T.~Shi, Y.~Li, H.~Xu, K.~Wu, G.~Qian, Q.~Chen \emph{et~al.}, ``Enhancing code llms with reinforcement learning in code generation,'' \emph{arXiv preprint arXiv:2412.20367}, 2024.

\bibitem{10628639}
Y.~Ji, Z.~Yu, and Y.~Wang, ``Assertion detection in clinical natural language processing using large language models,'' in \emph{2024 IEEE 12th International Conference on Healthcare Informatics (ICHI)}, 2024, pp. 242--247.

\bibitem{yang2025research}
Z.~Yang, Y.~Jin, Y.~Zhang, J.~Liu, and X.~Xu, ``Research on large language model cross-cloud privacy protection and collaborative training based on federated learning,'' \emph{arXiv preprint arXiv:2503.12226}, 2025.

\bibitem{li2025visual}
Y.~Li, Z.~Lai, W.~Bao, Z.~Tan, A.~Dao, K.~Sui, J.~Shen, D.~Liu, H.~Liu, and Y.~Kong, ``Visual large language models for generalized and specialized applications,'' \emph{arXiv preprint arXiv:2501.02765}, 2025.

\bibitem{li2024segmentation}
P.~Li, Y.~Lin, and E.~Schultz-Fellenz, ``Contextual hourglass network for semantic segmentation of high resolution aerial imagery,'' in \emph{2024 5th International Conference on Electronic Communication and Artificial Intelligence (ICECAI)}.\hskip 1em plus 0.5em minus 0.4em\relax IEEE, 2024, pp. 15--18.

\bibitem{jin2025adaptive}
Y.~Jin, Z.~Yang, X.~Xu, Y.~Zhang, and S.~Ji, ``Adaptive fault tolerance mechanisms of large language models in cloud computing environments,'' \emph{arXiv preprint arXiv:2503.12228}, 2025.

\bibitem{liu2024mt2st}
D.~Liu and Y.~Yu, ``Mt2st: Adaptive multi-task to single-task learning,'' \emph{arXiv preprint arXiv:2406.18038}, 2024.

\bibitem{xu2024empowering}
Y.~Xu and K.~Wu, ``Empowering developers: Ai-infused cloud services for software engineering,'' \emph{Asian American Research Letters Journal}, vol.~1, no.~1, 2024.

\bibitem{hrusto2024autonomous}
A.~Hrusto, P.~Runeson, and M.~C. Ohlsson, ``Autonomous monitors for detecting failures early and reporting interpretable alerts in cloud operations,'' in \emph{Proceedings of the 46th International Conference on Software Engineering: Software Engineering in Practice}, 2024, pp. 47--57.

\bibitem{stutz2024enhancing}
D.~Stutz, J.~T. de~Assis, A.~A. Laghari, A.~A. Khan, N.~Andreopoulos, A.~Terziev, A.~Deshpande, D.~Kulkarni, and E.~G. Grata, ``Enhancing security in cloud computing using artificial intelligence (ai),'' \emph{Applying Artificial Intelligence in Cybersecurity Analytics and Cyber Threat Detection}, pp. 179--220, 2024.

\bibitem{hassan2025managing}
N.~A.~B. Hassan, ``Managing data dependencies in cloud-based big data pipelines: Challenges, solutions, and performance optimization strategies,'' \emph{Orient Journal of Emerging Paradigms in Artificial Intelligence and Autonomous Systems}, vol.~15, no.~2, pp. 20--28, 2025.

\bibitem{tupe2025ai}
V.~Tupe and S.~Thube, ``Ai agentic workflows and enterprise apis: Adapting api architectures for the age of ai agents,'' \emph{arXiv preprint arXiv:2502.17443}, 2025.

\bibitem{chen2024transforming}
D.~Chen, A.~Youssef, R.~Pendse, A.~Schleife, B.~K. Clark, H.~Hamann, J.~He, T.~Laino, L.~Varshney, Y.~Wang \emph{et~al.}, ``Transforming the hybrid cloud for emerging ai workloads,'' \emph{arXiv preprint arXiv:2411.13239}, 2024.

\bibitem{cheng2023logai}
Q.~Cheng, A.~Saha, W.~Yang, C.~Liu, D.~Sahoo, and S.~Hoi, ``Logai: A library for log analytics and intelligence,'' \emph{arXiv preprint arXiv:2301.13415}, 2023.

\bibitem{tadi2022architecting}
S.~Tadi, ``Architecting resilient cloud-native apis: Autonomous fault recovery in event-driven microservices ecosystems,'' \emph{Journal of Scientific and Engineering Research}, vol.~9, no.~3, pp. 293--305, 2022.

\end{thebibliography}
}

\end{document}